\documentclass[conference]{IEEEtran}
\usepackage{times}
\usepackage[numbers]{natbib}
\usepackage{multicol}
\usepackage[bookmarks=true]{hyperref}
\usepackage{graphicx}
\usepackage{amsmath}
\usepackage{array 	}
\graphicspath{{pics/}}

\begin{document}

\title{Inferring 3D Articulated Models for Box Packaging Robot}


\author{Heran Yang, Tiffany Low, Matthew Cong and Ashutosh Saxena\\
Computer Science Department, Cornell University\\
\{hy279,twl46,mdc238\}@cornell.edu, asaxena@cs.cornell.edu}

\maketitle

\begin{abstract}
Given a point cloud, we consider inferring kinematic models of 3D articulated objects such as boxes for the purpose of manipulating them. While previous work has shown how to extract a planar kinematic model (often represented as a linear chain), such planar models do not apply to 3D objects that are composed of segments often linked to the other segments in cyclic configurations.
We present an approach for building a model that captures the relation between the input point cloud features and the object segment as well as the relation between the neighboring object segments. We use a conditional random field that allows us to model the dependencies between different segments of the object. 
We test our approach on inferring the kinematic structure from partial and
noisy point cloud data for a wide variety of boxes including cake boxes, pizza boxes, and cardboard cartons
of several sizes. The inferred structure enables our robot to successfully close these boxes by manipulating the
flaps.

\end{abstract}

\IEEEpeerreviewmaketitle

\section{Introduction}
Given a point cloud of a scene, we present a method for extracting an articulated 3D model that represents the kinematic structure of an object such as a box. We apply this to enable 
a robot to autonomously close boxes of several shapes and sizes. Such an ability is of interest to a personal
assistant robot as well as to commercial robots in applications such as packaging and shipping. 


Previous work on articulated structures was able to represent planar kinematic models as linear structures such as a chain of rigid bodies connected by joints. Linear structures greatly simplify the inference problem because they decompose the joint inference problem into independent sub-problems. Katz et al. 
\cite{Brock1,Brock2} considered linear articulated structures in 2D by relying on active vision techniques to learn kinematic properties of objects. 

\begin{figure}
\centering
\includegraphics[width=8cm]{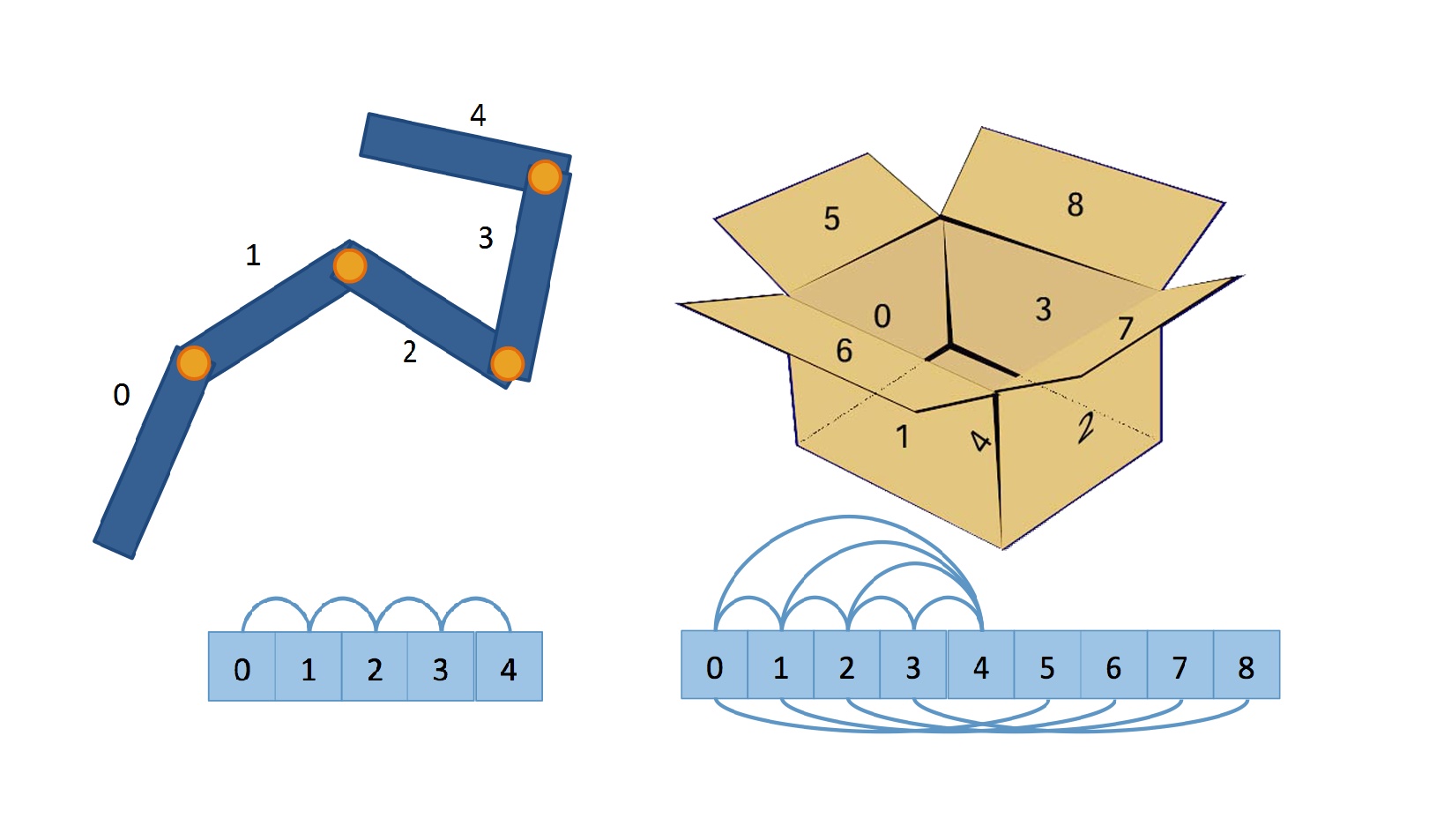}\vskip -2em
\caption{
\label{fig:boxmodel} An object consists of several segments. For planar objects (left), they can often be modeled as a chain
which involves very few relations (shown by curved lines on the bottom).  For 3D objects such as a box (right), 
the relation between different segments is more complicated.}
\end{figure}

However, complex objects cannot be easily represented by linear chains. For example, a box is an example of an oriented arrangement of segments which are highly interconnected. In a standard box (see Fig.~\ref{fig:boxmodel}), 
we can observe that the sides and base of the box are connected to four other faces while flaps extend outwards from only one face of the box. A linear model is unable to express these relations and constraints on the object's structure. Our proposed model for 3D kinematic objects allows for increased expressivity while maintaining computational tractability.

We present an approach for building such a 3D articulated model that captures the relation between 
the input point cloud features
the object segment as well as the relation between the neighboring object segments. 
We use an conditional random field (CRF) as an undirected graphical model that allows us to model the different segments of the object independently.


We evaluated our algorithm on boxes of varying sizes and structures over multiple experiments. We perceived the
scene (that often contains multiple boxes and other clutter) using a Microsoft Kinect camera.  The boxes were 
oriented at various angles causing a varying number of box flaps to be obscured. We obtain an overall accuracy of 76.55\% in building the box model and of 86.74\% in identifying enough planes to close the box. 
We have verified the robot closing several boxes in simulation and have also applied
our method on a robotic arm closing a box.

\section{Related Work}


Katz et al.~\cite{Brock1,Brock2} developed a relational representation of kinematic structure. In their model, a chain structure is used to model the kinematic properties of objects. Previous work by Sturm et al.~\cite{ 
burgard2} models motion that cannot be described by a simple prismatic or revolute joint. They successfully predicted the motion of drawers and cabinet doors. However, these joints are modeled within a linear structure.

There has been extensive prior work on object recognition in 3D environments using the RANSAC algorithm~\cite{fischler1981random, schnabel2007efficient} and the Hough transform~\cite{ballard1981generalizing}. These methods search through object primitives from the input image data in order to identify complex objects and have been successfully applied in performing a variety of manipulation tasks. For example, Rusu et al.~\cite{rusu2008towards} identified planes in a household kitchen environment which were fitted to models of common kitchen objects such as cupboards and tables.

The field of computer vision also contains some related work on part-based models involving the decomposition of objects into sub-parts. For example, Crandall et al.~\cite{HLtree} used a kinematic tree model to model human motion. This reduces the complexity of the search space \cite{HLprior} but maintains the representational power of the kinematic structure obtained. Hahnel et al.~\cite{hahnel2003learning} produced accurate models of indoor and outdoor environments that compare favorably to other methods which decompose and approximate environments using flat surfaces.
Other related works also use RGB-D data for different purposes in robot manipulation 
such as grasping \cite{jiang_grasping,aaai_grasping}, placing objects \cite{jiang_placing} and
human activity detection \cite{sung_activity}, where the point cloud data is used together with learning algorithms for the respective tasks.


\section{Our Approach}

The robot requires a good estimate of the box configuration before it can plan and execute a set of motions to close the box. The robot has to recognize boxes from the input point cloud data (see Fig.~\ref{fig:modelresults}). The task is challenging because we often have multiple boxes (along with other clutter) in the scene and because we consider a wide variety of boxes. In addition, the boxes are in random orientations and positions within the user environment.

An object is composed of several segments.
For example, a box can be described as a set of structured planes. We first segment the point cloud into clusters,  each of which is then decomposed into segments. Our goal is to learn the kinematic model $\mathcal{G}$ by correctly identifying the segments. In our application, the model is a box consisting of four sides, a base, and four flaps. Our goal is to find the optimal model $\mathcal{G}^*  = \arg \max \limits_{\mathcal{G}}J(\mathcal{G})$ with respect to a scoring function $J(\mathcal{G})$.

\subsection{Conditional Random Field (CRF)}
The process of identifying the different segments of 3D articulated structures is highly sequential and conditional. For instance, the position and orientation of a segment of a 3D object are likely to be well defined after a connected or nearby segment has been located. The relations we introduce in Section \ref{features} provide further examples. We therefore use a conditional random field (CRF) for the undirected probabilistic graphical model to determine the 3D articulated structure. 

Formally, we define $\mathcal{G} = (\mathcal{V},\mathcal{E})$ to be an undirected graph such that there is a node $v \in \mathcal{V}$ corresponding to each of the random variables that represents a segment in the 3D structure. Each edge $(v_i, v_j) \in \mathcal{E}$ represents that nodes $v_i$ and $v_j$ are not independent while a binary potential function, designated as $\Gamma(v_i, v_j)$, describes the relation or dependency between $v_i$ and $v_j$. Moreover, in some cases, segments of 3D articulated structures have internal features that are not dependent on any part of the object itself such as the natural orientations or natural sizes. Therefore, each node $v \in \mathcal{V}$ is associated with some unary potential functions and that we will refer to as $\phi(v_i)$. 

\subsection{Score Function for 3D Structure Modeling}

The scoring function $J(\mathcal{G})$ is a measure of the fitness of $\mathcal{G}$ over a set of features $\phi$ and relations $\Gamma$.  The corresponding weights are $w_{\phi}$ and $w_{\Gamma}$ for $\phi$ and $\Gamma$ respectively. The detailed explanations of $\phi$ and $\Gamma$ are in Section \ref{features} and the learning algorithm for the weights is covered in Section \ref{learning}.

\begin{equation} \label{eq:cost}
J(\mathcal{G}) = \sum\limits_{i \in \mathcal{V}}  w_\phi^T  \phi(v_i) +    \sum\limits_{(i,j) \in E} w_\Gamma^T \Gamma(v_i,v_j)
\end{equation}

\begin{figure}[here]
\centering
\includegraphics[width=3cm]{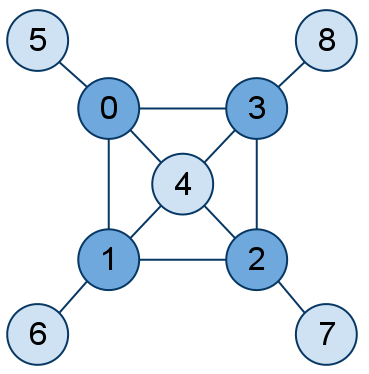}\vskip -1em
\caption{
\label{fig:graph} A graphical model of an articulated 3D object. Each node represents a segment of the object and every edge denotes a relation between two segments of the object. The figure presents an example of a box model with four flaps. Nodes 0, 1, 2 and 3 represent the sides of a box, node 4 represents the bottom of the box and nodes 5, 6, 7 and 8 represents the flaps of the box.  The edges of the graph represent connected faces of the box. 
}\vskip -1em
\end{figure}

\subsection{Complexity Analysis}\label{complexity}

An exhaustive search over all matchings of $n$ segments to a model consisting of $m$ segments is combinatorial to the order of $O(\frac{n!}{(n-m)!}) = O(n^m)$. Allowing for empty planes to be matched to the model segments does not increase the order of complexity $O(n^m)$. We show the state space can be significantly reduced in size if the segments are conditionally independent of each other. 

Each node $v_i$ has a set $\Delta_i$ of adjacent nodes. $v_i$ is independent of all nodes outside $\Delta_i$. When we are searching over the different matchings of the model and assigning possible segments to one node $v$, all the other independent nodes can be ignored due to conditional independence. In other words, if there are $t_i$ segments that have not yet been assigned and there are $k_i$ nodes $\in \Delta_i$ that have not yet been determined, we can restrict the search space to be $O(\frac{t_i!}{k_i!})$ for node $v_i$.

For the box model in Fig.~\ref{fig:graph}, the problem of matching segments to the flaps and the bottom face becomes a series of problems linear to the size of $n$ if the four sides have been determined. For example, in Fig.~\ref{fig:graph}, if the side plane nodes 0, 1, 2 and 3 are pre-chosen, the search space is $O(5(n-4)) = O(n)$, giving an overall state space of $O(n^4)O(n) = O(n^5)$ which is tractable for application problems such as ours.

\subsection{Feature and Relation Sets}\label{features}

For any collection of objects in a manipulation task, a set of features and relations can be chosen to distinguish them from other objects in the environment. A feature describes the segment relative to either a ground reference or the segment's properties. In comparison, a relation encodes the relative values of a set of properties between at least two segments within the model. Both features and relations require the tolerance parameter $\tau$ to define the bound of correctness. $\tau$ is defined as an angle/distance when it is used to bound orientation/location.
In our model, we have the following features and relations:

\emph{$\phi_1$: Absolute Orientation:} A majority of object models have a natural orientation, e.g. windows are vertical. $\phi_1$ measures the orientation of segment $i$ and compares it with unit vector $u$.   $\phi_1 = \tau - \| \arccos ( \frac {v_i^T u} {\| v_i \| } ) \| $. 

\emph{$\phi_2$: Absolute Location:} $\phi_2$ measures the closeness of the object model's segment $i$ to a reference point $L$. $\phi_2 = \tau - \| v_i - L \|$. 

\emph{$\phi_3$: Existence:} For real world data, it is hard for robots to distinguish between corrupted data and partially missing data from observation. $\phi_3$ assigns a reward for finding segment $i$ in the model. 

\emph{$\Gamma_1$: Relative Orientation:} Nearly all objects have rigid segments with fixed relative orientations, e.g. legs of tables are always perpendicular to their surfaces. $\Gamma_1$ compares the difference of the orientations of segment $i$ and $j$ to a specified angle $\alpha$. $\Gamma_1 = \tau - \| \alpha - \| \arccos ( \frac {v_i^T v_j} {\| v_i \cdot v_j \| } ) \| \|$. In the box model, connected sides and bottoms are perpendicular to each other and therefore $\alpha = \pi / 2 $.

\emph{$\Gamma_2$: Relative Location:} Nearly all objects have rigid segments with fixed relative position. $\Gamma_2$ compares the difference of the locations of segment $i$ and $j$ to a specified unit vector $u$. $\Gamma_2 = \tau - \| \arccos ( \frac {(v_i - v_j)^T u} {\| v_i - v_j \| } ) \| $. In the box model, flaps are always higher than sides and sides are always higher than base.

\emph{$\Gamma_3$: Segment Connectivity:} For objects with rotatable segments, $\Gamma_3$ measures the connectivity of the object model's segments $i$ and $j$ by calculating the distance in between. $\Gamma_2 = \tau - \| v_i - v_j \|$. In the box model, the sides, base, and flaps are all under such relation.

For complicated or obscure box models, more edges have to be added in addition to the base features and relations in Fig.~\ref{fig:graph}. For example, in the box model, a new relation $\gamma_4$ that measures the model's rectangular structure is later demonstrated via learning to be significantly crucial. For each two side planes $i, j$ that are across to each other, we find the four pairs of associated points $(p_i, p_j)$ and test if the vector $p_i - p_j$ is parallel to the side planes' orientations. In addition, $\gamma_5$ measures if two side planes that are across from each other are parallel. $\gamma_4$ and $\gamma_5$ combined give a satisfactory evaluation of the model's rectangular structure. However, $\gamma_4$ and $\gamma_5$ break the original conditional independence property which significantly increases the size of the search space. 


 \noindent

\subsection{Learning}\label{learning}
We collected ground-truth labeled data for training the parameters $w_\phi$ and $w_\gamma$. Features $\phi$ and $\gamma$ are computed from the box model $\mathcal{G}$ that our algorithm built from the point cloud. For each pair of $\phi$ and $\gamma$, we obtain $J(\mathcal{G})$ and check the correctness of $\mathcal{G}$ by comparing it with the labeled data, i.e. the ratio of correctly marked planes to the total number of planes in the labeled model. Note that the equation~\ref{eq:cost} is linear in $w_\phi$ and $w_\gamma$. Therefore, we can use the normal equations with regularization to estimate the parameters \cite{hastie2001elements}.

\subsection{Inference}
Following the conditional independence assumption encoded by the graphical model in Fig.~\ref{fig:graph} (see Section~\ref{complexity}), we only have to consider a tractable number of box assignments for inferring the optimal configuration. We evaluate scores  for all cases and select the optimal one. In practice, we evaluate
the ones with higher $\phi$'s by forming a priority queue which speeds up the inference further.


However, for more complex models, i.e. too many relations in the graph, using the full relation set may cause no independent segments. In this case, we remove certain relations that are determined by learning to be the least important and produce a sparse graph with fewer edges. Then we apply the above procedure to the simple graph and receive the updated list $L$ of all candidate matchings. We then selected the top $c\|L\|$ elements of the candidate matchings and put them in a priority queue that ranks by applying the full feature set to $\mathcal{G}$. The top element of the resulting queue is chosen as the optimal predicted solution $\hat{\mathcal{G}^*}$. Variable $c$ depends on the computational power and we found from experiments that as $c$ becomes larger than $1/4$, the optimal solution converges to the predicted $\hat{\mathcal{G}^*}$.

\subsection{Segment Identification}
Given the point cloud data, segments need to be extracted from the main clusters and be correctly identified to build representative object models,. We apply the RANSAC algorithm \cite{fischler1981random} to extract segments from the point cloud. In our case, RANSAC is applied iteratively and the best fitting planes to the clusters are also obtained sequentially. However the planes returned from RANSAC are not bounded by rectangles and may contain outliers. We therefore fetch the extracted planes by filtering out outliers using Euclidean clustering. To obtain the corners from the point cloud associated with each plane, we construct the convex hulls of the points to eliminate redundant points and find the minimum bounding rectangles of the planes. 

By going through the above procedures, planes can be obtained by almost 100\% accuracy. However, any noise in the environment that is attached to the main cluster can also potentially be identified as planes which may confuse the searching algorithm. Therefore we ignore the noise if the error rate evaluated from RANSAC is over a pre-defined limit. We also rely on the feature/relation performance for boxes with noise attached. 


\section{Planning and Control}

To close a box in the environment, the robot needs to identify and locate the box to manipulate it into the desired state. Each flap can be closed from a set of paths. Given the box model with the location and orientation of each flap, we use OpenRAVE's \cite{diankov2008openrave} under-constrained inverse kinematics solver. The program applies brute-force search approach by defining intermediate rotating planes. The planner will pick sampled points from each intermediate planes and search for paths through them. 

To decide the order of closing the flaps, our program sorts the flaps in order of ascending area, which are given by the box models. Then the manipulator will greedily close the flaps in that order. For the three flaps that are closer to the robot arm, the planner is able to find valid paths in 90\% of the experiments. However for the farthest flap, the planner only has a success rate of 50\% due to the robot's limited workspace.

\begin{figure*}
\begin{center}

\includegraphics[height=2.5cm, width=2.75cm]{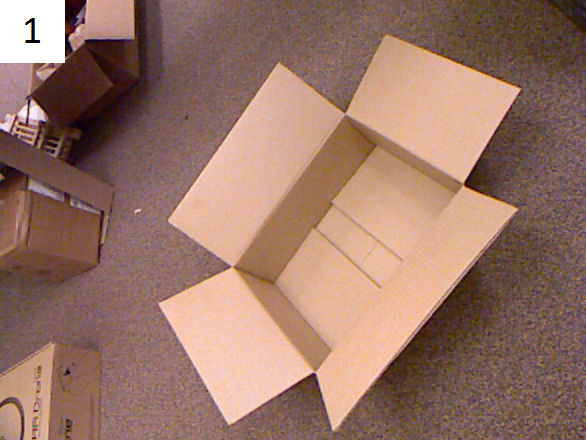}
\includegraphics[height=2.5cm, width=2.75cm]{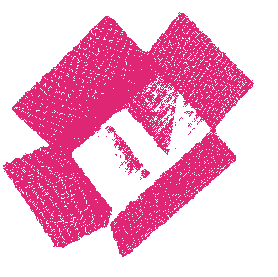}
\includegraphics[height=2.5cm, width=2.75cm]{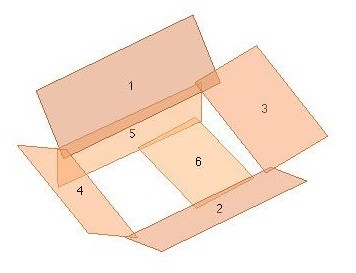}
\includegraphics[height=2.5cm, width=2.75cm]{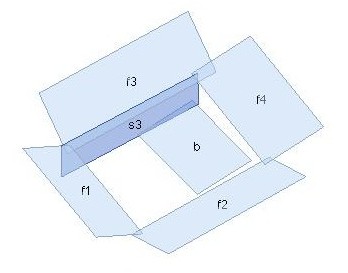} 
\includegraphics[height=2.5cm, width=2.5cm]{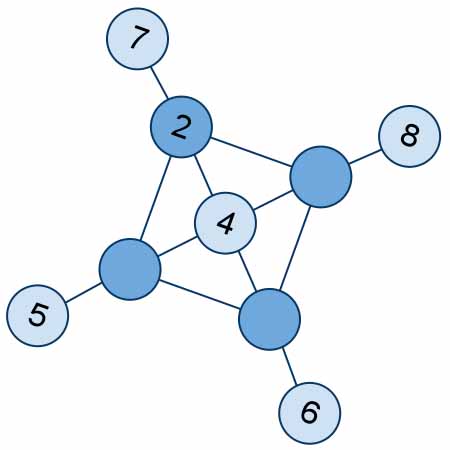} 

\includegraphics[height=2.5cm, width=2.75cm]{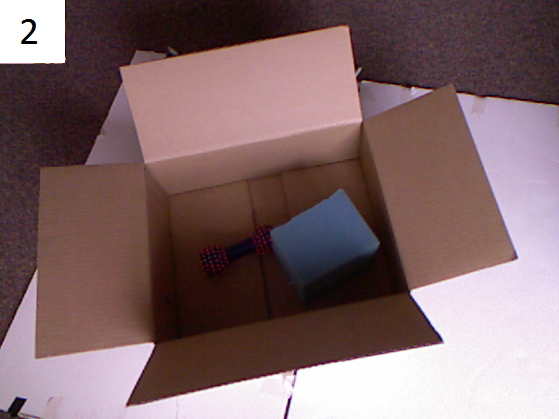} 
\includegraphics[height=2.5cm, width=2.75cm]{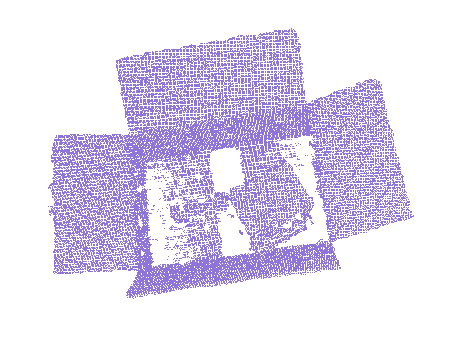}  
\includegraphics[height=2.5cm, width=2.75cm]{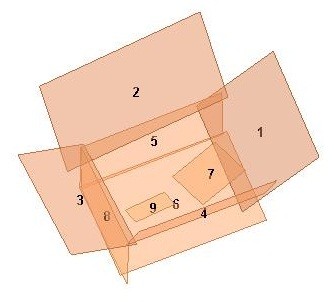} 
\includegraphics[height=2.5cm, width=2.75cm]{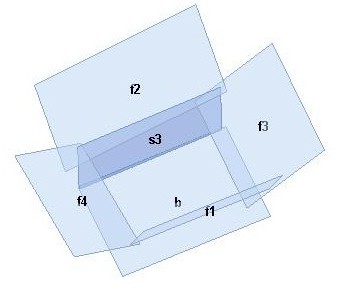} 
\includegraphics[height=2.5cm, width=2.5cm]{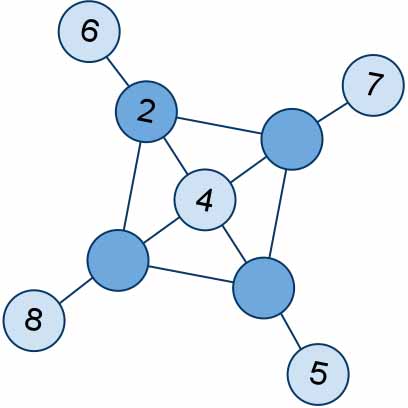} 

\includegraphics[height=2.5cm, width=2.75cm]{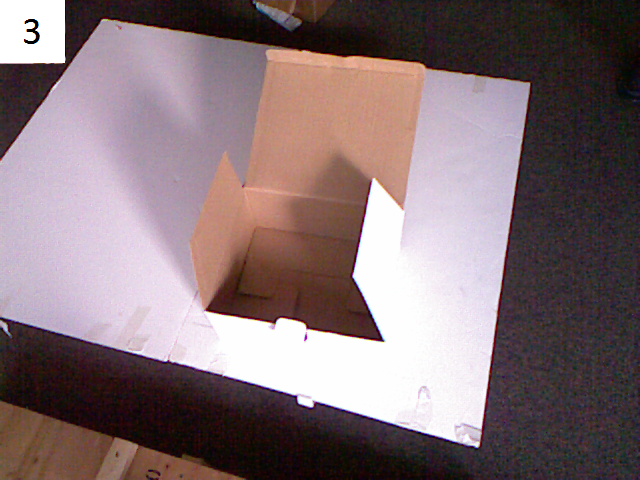}
\includegraphics[height=2.5cm, width=2.75cm]{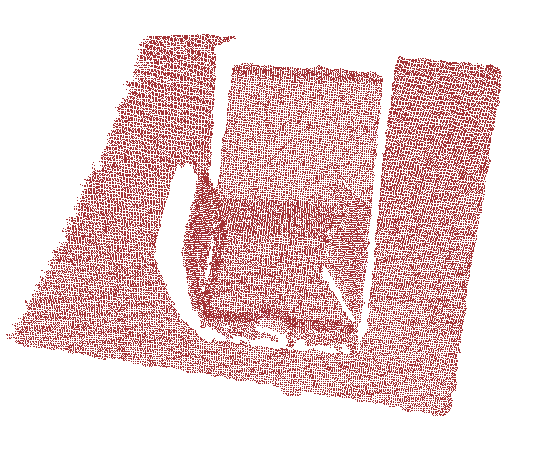} 
\includegraphics[height=2.5cm, width=2.75cm]{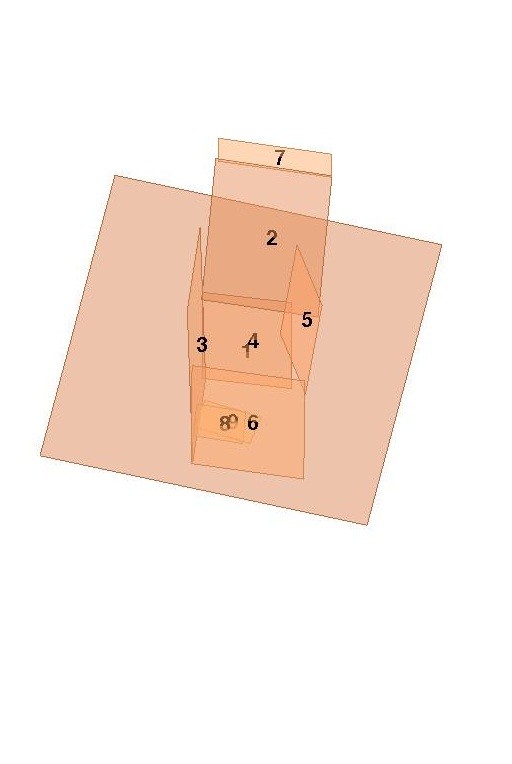}
\includegraphics[height=2.5cm, width=2.75cm]{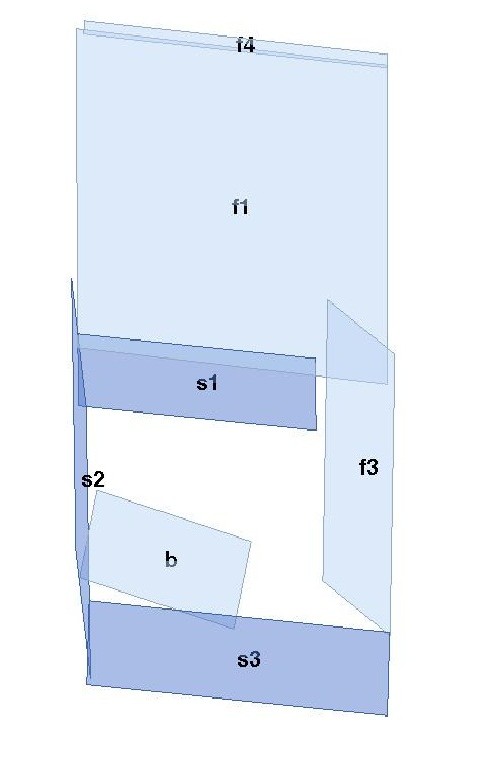}
\includegraphics[height=2.5cm, width=2.5cm]{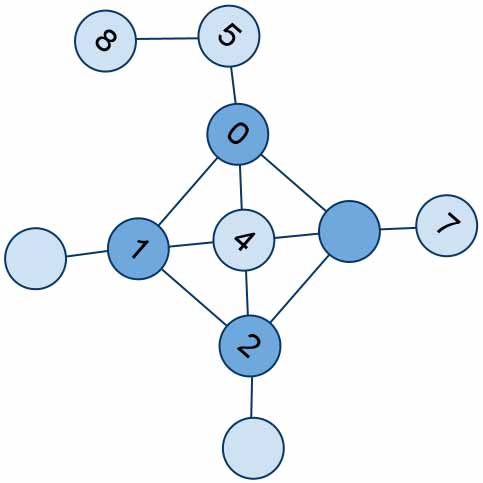}

\includegraphics[height=2.5cm, width=2.75cm]{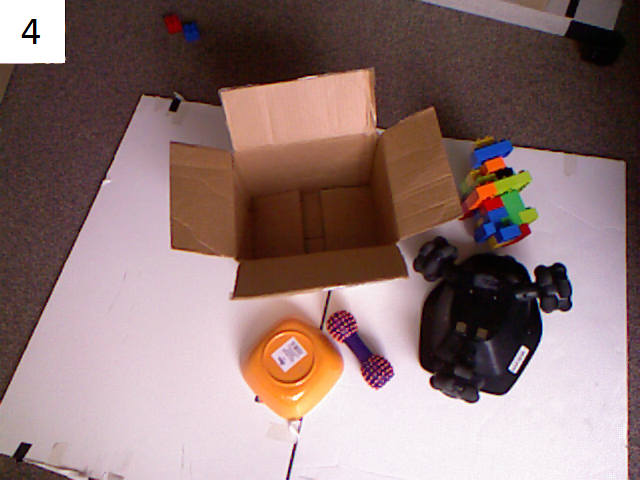} 
\includegraphics[height=2.5cm, width=2.75cm]{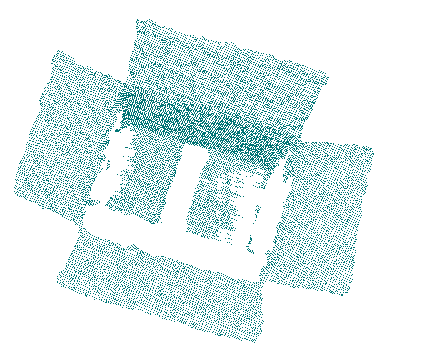}  
\includegraphics[height=2.5cm, width=2.75cm]{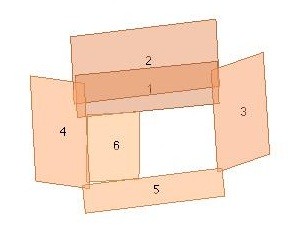} 
\includegraphics[height=2.5cm, width=2.75cm]{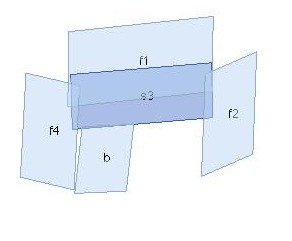} 
\includegraphics[height=2.5cm, width=2.5cm]{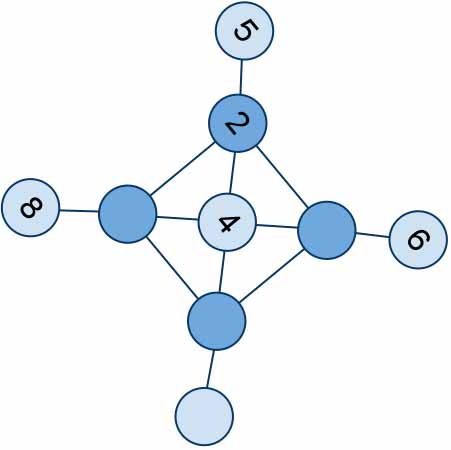} 

\includegraphics[height=2.5cm, width=2.75cm]{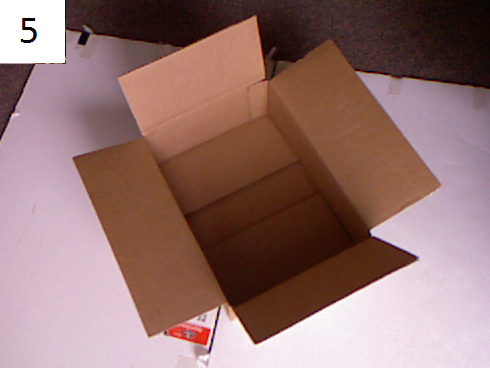}
\includegraphics[height=2.5cm, width=2.75cm]{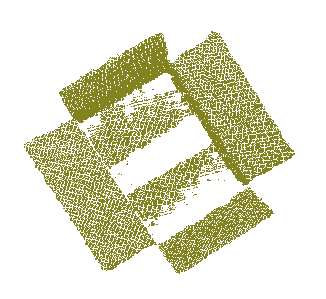} 
\includegraphics[height=2.5cm, width=2.75cm]{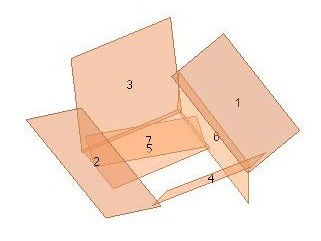}
\includegraphics[height=2.5cm, width=2.75cm]{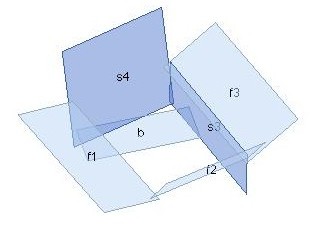}
\includegraphics[height=2.5cm, width=2.5cm]{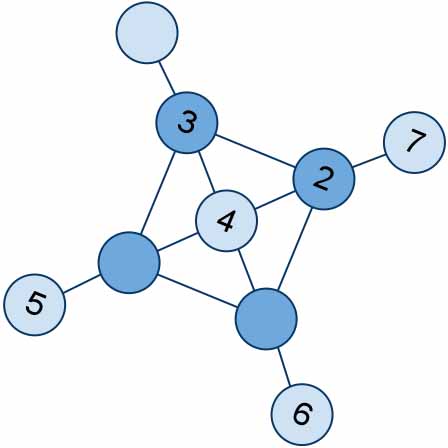}

\includegraphics[height=2.5cm, width=2.75cm]{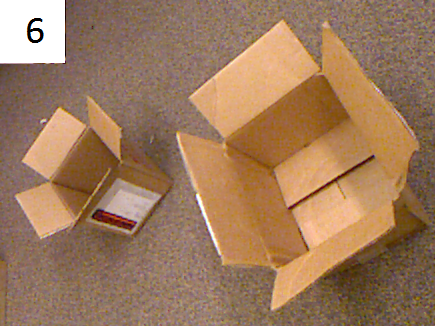}
\includegraphics[height=2.5cm, width=2.75cm]{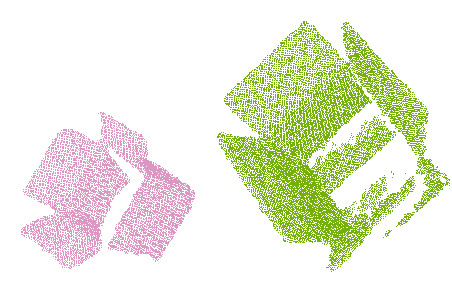} 
\includegraphics[height=2.5cm, width=2.75cm]{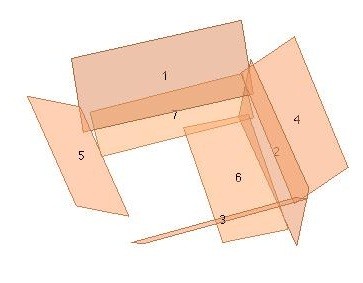}
\includegraphics[height=2.5cm, width=2.75cm]{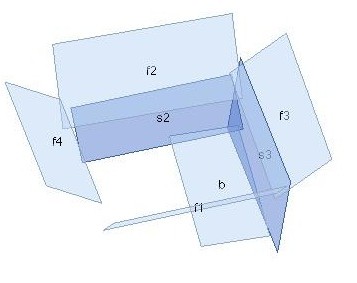}
\includegraphics[height=2.5cm, width=2.5cm]{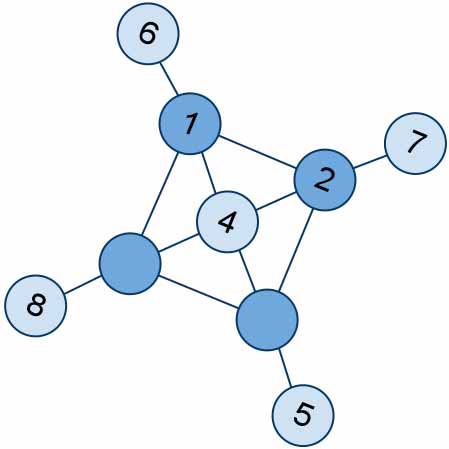}

\includegraphics[height=2.5cm, width=2.75cm]{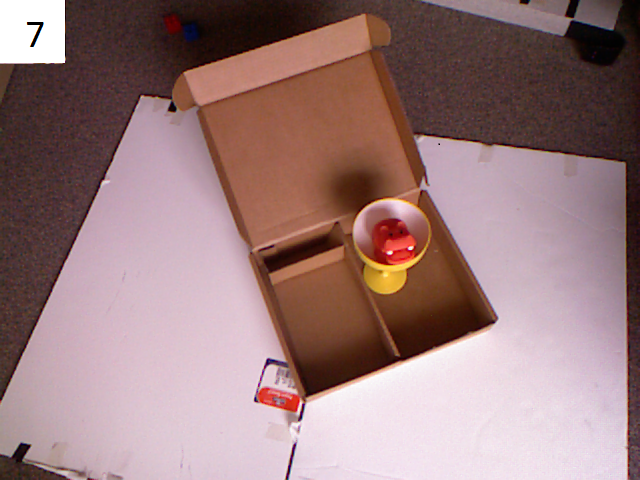}
\includegraphics[height=2.5cm, width=2.75cm]{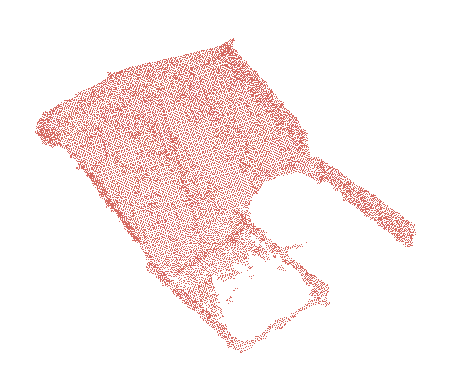} 
\includegraphics[height=2.5cm, width=2.75cm]{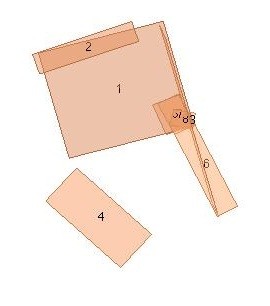}
\includegraphics[height=2.5cm, width=2.75cm]{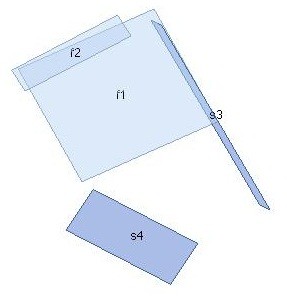}
\includegraphics[height=2.5cm, width=2.5cm]{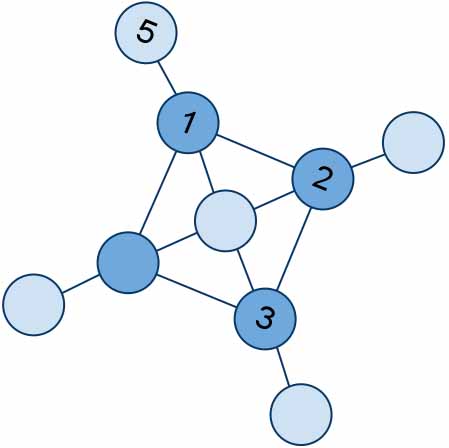}

\includegraphics[height=2.5cm, width=2.75cm]{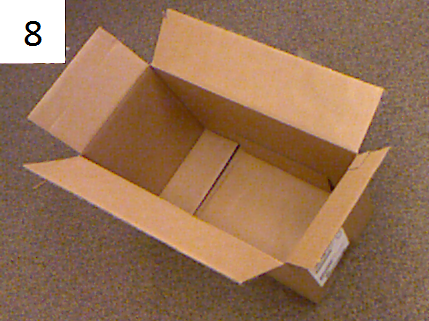}
\includegraphics[height=2.5cm, width=2.75cm]{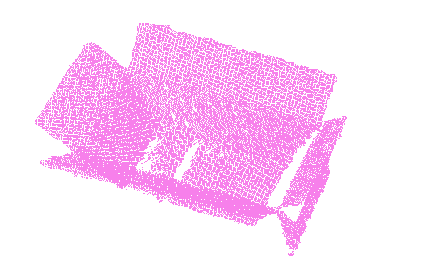} 
\includegraphics[height=2.5cm, width=2.75cm]{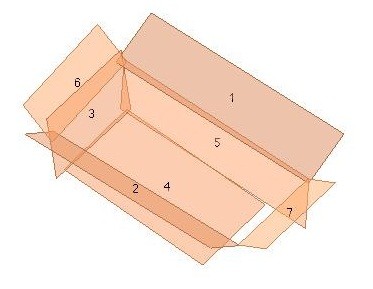}
\includegraphics[height=2.5cm, width=2.75cm]{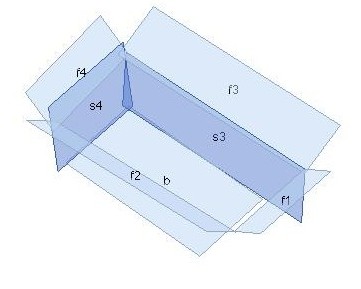}
\includegraphics[height=2.5cm, width=2.5cm]{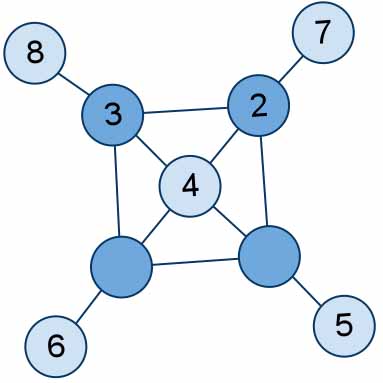}

 %
 %
 %
 %
 %
 %
 %

%

\caption{
\label{fig:modelresults}The process of inferring box models from point cloud data: We first begin with Kinect RGBD data (first column from left) and extract out the ground plane. Then we segment planes from the point cloud cluster (second column) and find the minimum bounding rectangle (third column). Finally, we run the segments on the inference algorithm to return the best matching to the box model (fourth column). The darker shaded planes represent the side faces, while the lighter ones represent the flaps and the base of the box.}
\end{center}
\end{figure*}

\begin{figure}[tb!]
\centering
\parbox{8cm}{\centering 
\includegraphics[height=2.8cm,width=3.6cm]{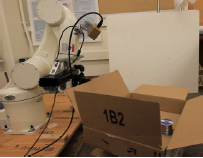} 
\includegraphics[height=2.8cm,width=3.6cm]{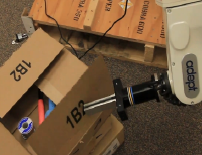}
\vskip 0.4em
\includegraphics[height=2.6cm,width=3.6cm]{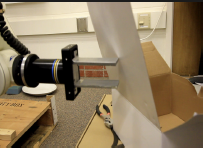} 
\includegraphics[height=2.6cm,width=3.6cm]{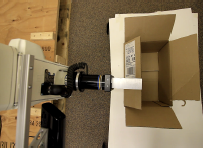}}

\caption{ 
\label{fig:robot} The robot arm with gripper closing a flap on a box. After identifying a box in the point cloud data, the robot executes a series of planned paths to close the flaps of the box.}
\end{figure}

\begin{figure}[here]
\centering
\includegraphics[width=0.30\columnwidth]{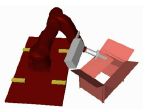}
\includegraphics[width=0.30\columnwidth]{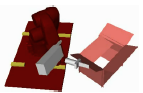}
\includegraphics[width=0.30\columnwidth]{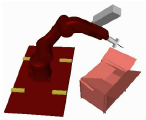}\vskip -1em
\caption{A simulation of the robot closing a box flap. We ensure that the path found is feasible, obstacle-free, and oriented in the correct direction to close the flap. The robot plans an arc from outside of the box to the inside of the box that intersects the flap.}\vskip -1em
\label{fig:robotexp}
\end{figure}

\section{Hardware}
Our experiments were conducted using an Adept Viper s850 6-DOF arm with a parallel-plate gripper giving a reach of approximately 100 centimeters. In order to obtain point cloud data, a Microsoft Kinect was mounted on the robot in a position (pictured in Fig.~\ref{fig:robot}) that allowed for changes in the orientation of the camera in order to obtain a variety of viewpoints.


\section{Experiments}

To demonstrate the robustness of our algorithm, we collected point cloud data for several classes of boxes including various sizes of standard packaging boxes and unusually shaped boxes (see Fig.~\ref{fig:modelresults}). We then ran a series of experiments on an extensive dataset to determine the accuracy of the inference algorithm and then verified that the robot was able to close the box through simulation. 

  \subsection{Data}

Our testing environment contains a total of 50 point clouds and has a total of 15 different types of boxes in the set. We relied on 12 of these point clouds for training purposes and the rest for testing. 

We considered a set of standard cardboard boxes which were sorted into three size categories and tested the algorithm on unusual boxes including cake boxes, pizza boxes and a cardboard carton. For each box type, we collected images of the box from different orientations. The point clouds in our dataset often included extraneous common household objects placed among the scene such as toys, dishes, and cups.  

\subsection{Accuracy Evaluation}
We measure flap inference accuracy and full model inference accuracy on our box model (see Table \ref{fig:table}). We define a model's flap inference accuracy as (the number of correctly identified flaps - the number of wrongly identified flaps) / (the number of flaps in the point cloud data). Similarly, the full model inference accuracy = (the number of correctly identified planes - the number of wrongly identified planes) / (the number of planes in the point cloud data). Notice that all four rotations of the box model are acceptable. For box closing, flap inference accuracy is the most important metric and we define it to be the percentage of correctly identified flap segments.

  \subsection{Results on the Dataset}

Fig.~\ref{fig:modelresults} shows the original image, the point cloud data, the segments observed, the matching returned by the algorithm, and finally the conditional random field graphical model. We see that the robot can identify a diverse set of boxes. The performance remains stable for boxes of various sizes and the algorithm is able to recognize non-standard boxes. This is seen in examples 3 and 7, where the structure of a flap on a flap is correctly identified despite its deviation from the labeled model.

Our algorithm is able to filter out noise in the data. For the second example, the segments belonging to items in the box are discarded in the matched model. Similarly, the algorithm is able to remove the sources of noise in the data for examples 4 and 7. When more than one box is in the scene, the algorithm is able to recognize the segments as individual boxes. This is seen in example 6.

\begin{table}
\caption{\label{fig:table}Performance of the inference algorithm on test point cloud inputs. }
\vskip -1em
\centering
{\small
\begin{tabular}{l|c|c}\hline

\hline
Box type     & \parbox[c][2.6em]{2.3cm}{\centering Flap Inference Accuracy (\%)} & \parbox[c][2.6em]{2.9cm}{\centering Full Model Inference Accuracy (\%)}\\\hline
Small boxes  & 82.50                 & 79.64\\ 
Medium boxes & 88.54                 & 76.00\\
Large boxes  & 90.00                 & 78.04\\
Cartons      & 87.50                 & 66.96\\
Cake boxes   & 85.00                 & 70.50\\
Pizza boxes  & 85.41                & 73.06\\ \hline
Full dataset & 86.74                 & 76.55\\ \hline

\hline
\end{tabular}}

\end{table}

The algorithm is able to identify box models with a 76.55\% accuracy. For the closing task, we only require the correct identification of flaps which we obtain with an accuracy of 86.74\% Therefore the robot shows good performance in identifying flaps to be closed. This also demonstrates that the algorithm is able to tolerate noise and different box configurations and types. 

Finally, we demonstrate that our algorithm and its associated articulated model can be applied to real-world tasks through a set of robotic experiments where the robot successfully closes the identified flaps.  Please see the video at:
\texttt{http://pr.cs.cornell.edu/articulated3d}





\section*{Acknowledgments}

We acknowledge Yun Jiang, Akram Helou, Marcus Lim and Stephen Moseson for useful
discussions.

\bibliographystyle{IEEEtran}
\bibliography{box_report_rss}

\vspace*{2in}

\end{document}